\pdfoutput=1

\documentclass[11pt]{article}

\usepackage[]{EMNLP2022}

\usepackage{times}
\usepackage{latexsym}

\usepackage[T1]{fontenc}

\usepackage[utf8]{inputenc}

\usepackage{microtype}

\usepackage{inconsolata}

\usepackage{mathrsfs}
\usepackage{amsmath}
\usepackage{graphicx}
\usepackage{float}
\usepackage{booktabs}
\usepackage{colortbl}
\usepackage{tabularx}
\usepackage{pbox}
\usepackage{amsfonts, amssymb}
\usepackage{cleveref}
\usepackage{hyperref}
\usepackage{makecell}
\usepackage{verbatim}
\usepackage{paralist}
\usepackage[usestackEOL]{stackengine}
\PassOptionsToPackage{table}{xcolor}

\usepackage{lipsum} 
\usepackage{enumitem}
\setlist{nosep} 

\usepackage{hyperref}

\usepackage[T1]{fontenc}

\usepackage[utf8]{inputenc}

\usepackage{microtype}

\DeclareMathOperator*{\argmax}{arg\,max}
\DeclareMathOperator{\sgn}{sgn}
\newcommand{\ts}{\textsuperscript}

\title{HashFormers: Towards Vocabulary-independent Pre-trained Transformers}

\author{Huiyin Xue \and Nikolaos Aletras \\
        Department of Computer Science, University of Sheffield \\
United Kingdom \\
 \texttt{\{hxue12, n.aletras\}@sheffield.ac.uk}}

\begin{document}
\maketitle

\begin{abstract}
Transformer-based pre-trained language models are vocabulary-dependent, mapping by default each token to its corresponding embedding. This one-to-one mapping results into embedding matrices that occupy a lot of memory (i.e. millions of parameters) and grow linearly with the size of the vocabulary. Previous work on on-device transformers dynamically generate token embeddings on-the-fly without embedding matrices using locality-sensitive hashing over morphological information. These embeddings are subsequently fed into transformer layers for text classification. However, these methods are not pre-trained. Inspired by this line of work, we propose \textsc{HashFormers}, a new family of vocabulary-independent pre-trained transformers that support an unlimited vocabulary (i.e. all possible tokens in a corpus) given a substantially smaller fixed-sized embedding matrix. We achieve this by first introducing computationally cheap hashing functions that bucket together individual tokens to embeddings. We also propose three variants that do not require an embedding matrix at all, further reducing the memory requirements. We empirically demonstrate that \textsc{HashFormers} are more memory efficient compared to standard pre-trained transformers while achieving comparable predictive performance when fine-tuned on multiple text classification tasks. For example, our most efficient \textsc{HashFormer} variant has a negligible performance degradation (0.4\% on GLUE) using only 99.1K parameters for representing the embeddings compared to 12.3-38M parameters of state-of-the-art models.\footnote{Code is available here: \url{https://github.com/HUIYINXUE/hashformer} and the pre-traied HashFormer models are available here: \url{https://huggingface.co/klein9692}.}

\end{abstract}

\section{Introduction}

The majority of transformer-based~\citep{vaswani2017attention} pre-trained language models~(PLMs; \citealt{devlin-etal-2019-bert,liu2019roberta,dai-etal-2019-transformer,yang2019xlnet}) are vocabulary-dependent, with each single token mapped to its corresponding vector in an embedding matrix. This one-to-one mapping makes it impractical to support out-of-vocabulary tokens such as misspellings or rare words~\citep{pruthi-etal-2019-combating,sun2020adv}. Moreover, it linearly increases the memory requirements with the vocabulary size for the token embedding matrix~\citep{chung2021rethinking}. For example, given a token embedding size of 768, \textsc{BERT-base} with a vocabulary of 30.5K tokens needs 23.4M out of 110M total parameters while \textsc{RoBERTa-base} with 50K tokens needs 38M out of 125M total parameters. Hence, disentangling the design of PLMs from the vocabulary size and tokenization approaches would inherently improve memory efficiency and pre-training, especially for researchers with access to limited computing resources~\citep{strubell2019energy,schwartz2020green}. 

Previous efforts for making transformer-based models vocabulary-independent include dynamically generating token embeddings on-the-fly without embedding matrices using hash embeddings~\citep{svenstrup2017hash,ravi2019efficient} over morphological information~\cite{sankar-etal-2021-device}. However, these embeddings are subsequently fed into transformer layers trained from scratch for on-device text classification without any pre-training. \citet{clark-et-al2022-canine} proposed CANINE, a model that operates on Unicode characters using a low-collision multi-hashing strategy to support \textasciitilde1.1M Unicode codepoints as well as infinite character four-grams. This makes CANINE independent of tokenization while limiting the parameters of its embedding layer to 12.3M. \citet{xue-etal-2022-byt5} proposed models that take as input byte sequences representing characters without explicit tokenization or a predefined vocabulary to pre-train transformers in multilingual settings.

In this paper, we propose \textsc{HashFormers} a new family of vocabulary-independent PLMs. Our models support an unlimited vocabulary (i.e. all possible tokens in a given pre-training corpus) with a considerably smaller fixed-sized embedding matrix. We achieve this by employing simple yet computationally efficient hashing functions that bucket together individual tokens to embeddings inspired by the hash embedding methods of \citet{svenstrup2017hash} and \citet{sankar-etal-2021-device}. Our contributions are as follows:
\begin{enumerate}
    \item To the best of our knowledge, this is the first attempt towards reducing memory requirements of PLMs using various hash embeddings with different hash strategies aiming to substantially reduce the embedding matrix compared to the vocabulary size;
    \item Three \textsc{HashFormer} variants further reduce the memory footprint by entirely removing the need of an embedding matrix;
    \item We empirically demonstrate that our \textsc{HashFormers} are consistently more memory efficient compared to vocabulary-dependent PLMs while achieving comparable predictive performance when fine-tuned on a battery of standard text classification tasks.
    
\end{enumerate}

\section{Related Work}

\subsection{Tokenization and Vocabulary-independent Transformers}
Typically, PLMs are pre-trained on text that has been tokenized using subword tokenization techniques such as WordPiece~\citep{wu2016google}, Byte-Pair-Encoding (BPE; \citealt{sennrich-etal-2016-neural}) and SentencePiece~\citep{kudo-richardson-2018-sentencepiece}.

Attempts to remove the dependency of PLMs on a separate tokenization component include models that directly operate on sequences of characters~\citep{tay2022charformer,el-boukkouri-etal-2020-characterbert}. However, these approaches do not remove the requirement of an embedding matrix. Recently, \citet{xue-etal-2022-byt5} proposed PLMs that take as input byte sequences representing characters without explicit tokenization or a predefined vocabulary in multilingual settings. PLMs in \citet{clark-et-al2022-canine} operating on Unicode characters or ngrams also achieved the similar goal. These methods improve memory efficiency but still rely on a complex process to encode the relatively long ngram sequences of extremely long byte/Unicode sequences, affecting their computational efficiency.

In a different direction, \citet{sankar-etal-2021-proformer} proposed \textsc{ProFormer}, an on-device vocabulary-independent transformer-based model. It generates token hash embeddings~\citep{svenstrup2017hash, shi2009hash,ganchev-dredze-2008-small} on-the-fly by applying locality-sensitive hashing over morphological features. Subsequently, hash embeddings are fed to transformer layers for text classification. However, \textsc{ProFormer} is trained from scratch using task-specific data without any pre-training.

\subsection{Compressing PLM Embeddings}

A different line of work has focused on compressing the embedding matrix in transformer models~\citep{ganesh-etal-2021-compressing}. \citet{prakash-etal-2020-compressing} proposed to use compositional code embeddings~\citep{shu2018compressing} to reduce the size of the embeddings in PLMs for semantic parsing. \citet{zhao-etal-2021-extremely} developed a distillation method to align teacher and student token embeddings using a mixed-vocabulary training (i.e. the student and teacher models have different vocabularies) for learning smaller \textsc{BERT} models. However, these approaches still rely on a predefined vocabulary. \citet{clark-et-al2022-canine} adopted low-collision multi-hashing strategy to support \textasciitilde1.1M Unicode codepoints and a larger space of character four-grams with a relatively small embedding matrix containing 12.3M parameters.


\section{HashFormers}\label{sec:HashFormers}
In this section, we present \textsc{HashFormers}, a family of vocabulary-independent hashing-based pre-trained transformers. 

\subsection{Many-to-One Mapping from Tokens to an Embedding}
\begin{figure}[!h]
    \centering
    \includegraphics[width=0.48\textwidth]{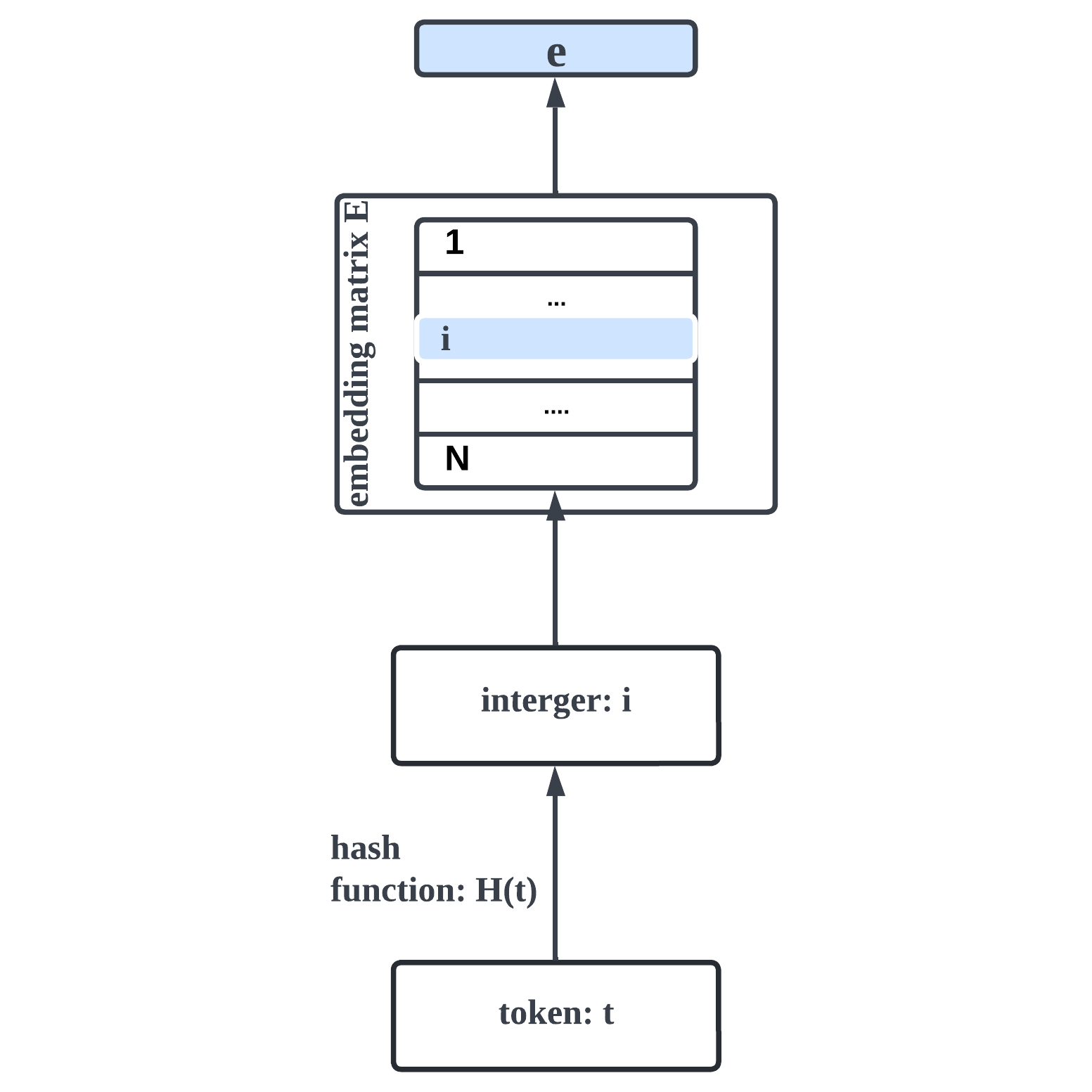}
    \caption{\textsc{HashFormer}-Emb.}
    \label{fig:fig-hformer-emb}
\end{figure}

Given a token $t$, \textsc{HashFormers} use a hash function $\mathcal{H}$ to map $t$ into a value $v$. Using hashing allows our model to map many tokens into a single embedding and support an infinite vocabulary. We obtain the embedding index by squashing its hash value $v$ into $i=[1,..., N]$ where $\mathbf{e}=\mathbf{E}_i$ is the corresponding embedding from a matrix $\mathbf{E} \in \mathbb{R}^{N\times d}$ where $N$ is the number of the embeddings and $d$ is their dimensionality. We assume that $|V|\gg N$ where $|V|$ is the size of the vocabulary. Subsequently, $\mathbf{e}$ is passed through a series of transformer layers for pre-training. This is our first variant, \textsc{HashFormer}-Emb that relies on a look-up embedding matrix (see Figure~\ref{fig:fig-hformer-emb}). Our method is independent of tokenization choices.

\subsection{Message-Direct Hashing (HashFormers-MD)} \label{sec:HashFormers-MD}

Our first approach to hash tokens is by using a Message-Digest (MD5) hash function~\citep{rivest1992md5} to map each token to its 128-bits output, $v=\mathcal{H}(t)$. The mapping can be reproduced given the same secret key. MD5 is a `random' hashing approach, returning mostly different hashes for tokens with morphological or semantic similarities. For example:
\begin{align*}
\small
 \text{MD5}(\text{`play'}) &= \textit{\small d077f244def8a70e5ea758bd8352fcd8}\\
 \small
 \text{MD5}(\text{`plays'}) &= \textit{\small 4a258d930b7d3409982d727ddbb4ba88}   
\end{align*}
\noindent It is simple and does not require any pre-processing to obtain the bit encoding for each token. To map the hash output $v$ into its corresponding embedding, we transform its binary value into decimal and then compute the index $i$ to $\mathbf{E}$ as $i=v\mathbin{\%}N$.

\subsection{Locality-Sensitive Hashing (\textsc{HashFormers-LSH})} \label{sec:HashFormers-LSH}
Locality-sensitive hashing (LSH) hashes similar tokens into the same indices with high probability~\citep{rajaraman2011mining}. \textsc{HashFormer-LSH} uses LSH hashing to assign tokens with similar morphology (e.g. `play', `plays', `played') to the same hash encoding. This requires an additional feature extraction step for token representation. 

\paragraph{Token to Morphological Feature Vector:}
We want to represent each token with a vector $\mathbf{x}$ as a bag of morphological (i.e. character n-grams) features. For each token, we first extract character n-grams ($n \in {1,2,3,4}$) to get a feature vector whose dimension is equal to the vocabulary size.\footnote{We keep the top-50K most frequent n-grams in the pre-training corpus.} Each element in the feature vector is weighted by the frequency of the character n-grams of the token.

\paragraph{Morphological Vector to Hash Index:}
Once we obtain the morphological feature vector of each token, we first define $N$ random hyperplanes, each represented by a random unit vector $\mathbf{r^i}\in\mathbb{R}^{d_x}$, where $d_x$ is the dimensionality of the morphological feature vector. 
Following a similar approach to~\citet{Kitaev2020Reformer:}, we compute the hash value $v$ as the index of the nearest random hyperplane vector to the token's feature vector, $\mathbf{x}$ obtained by computing $v=\mathcal{H}(\mathbf{x}) = \argmax(\mathbf{x}\mathbf{R}), \mathbf{R}=[\mathbf{r}^1;...;\mathbf{r}^{N}]$ where $[\alpha; \beta]$ denotes the concatenation of two vectors. This approach results into bucketing together  tokens with similar morphological vectors. Similar to \textsc{HashFormer-MD}-Emb, we compute the embedding index as $i=v\mathbin{\%}N$.

To prevent storing a large projection matrix ($\mathbb{R}^{d_x\times N}$) for accommodating each unit vector, we design an on-the-fly computational approach. We only store a vector $\mathbf{\eta} \in \mathbb{R}^{d_x}$ that is randomly initialized from the standard normal distribution, guaranteeing that each column $\mathbf{r}$ in the matrix $\mathbf{R}$ is a permutation of $\mathbf{\eta}$ with a unique offset value (e.g. $\mathbf{r}^1=[\mathbf{\eta}_2,...,\mathbf{\eta}_{N},\mathbf{\eta}_1]$). Each offset value only relies on the index of the hyperplane. This setting ensures that each hyperplane has the same L2-norm.

\subsection{Compressing the Embedding Space}
\begin{figure}[!htp]
    \centering
    \includegraphics[width=0.48\textwidth]{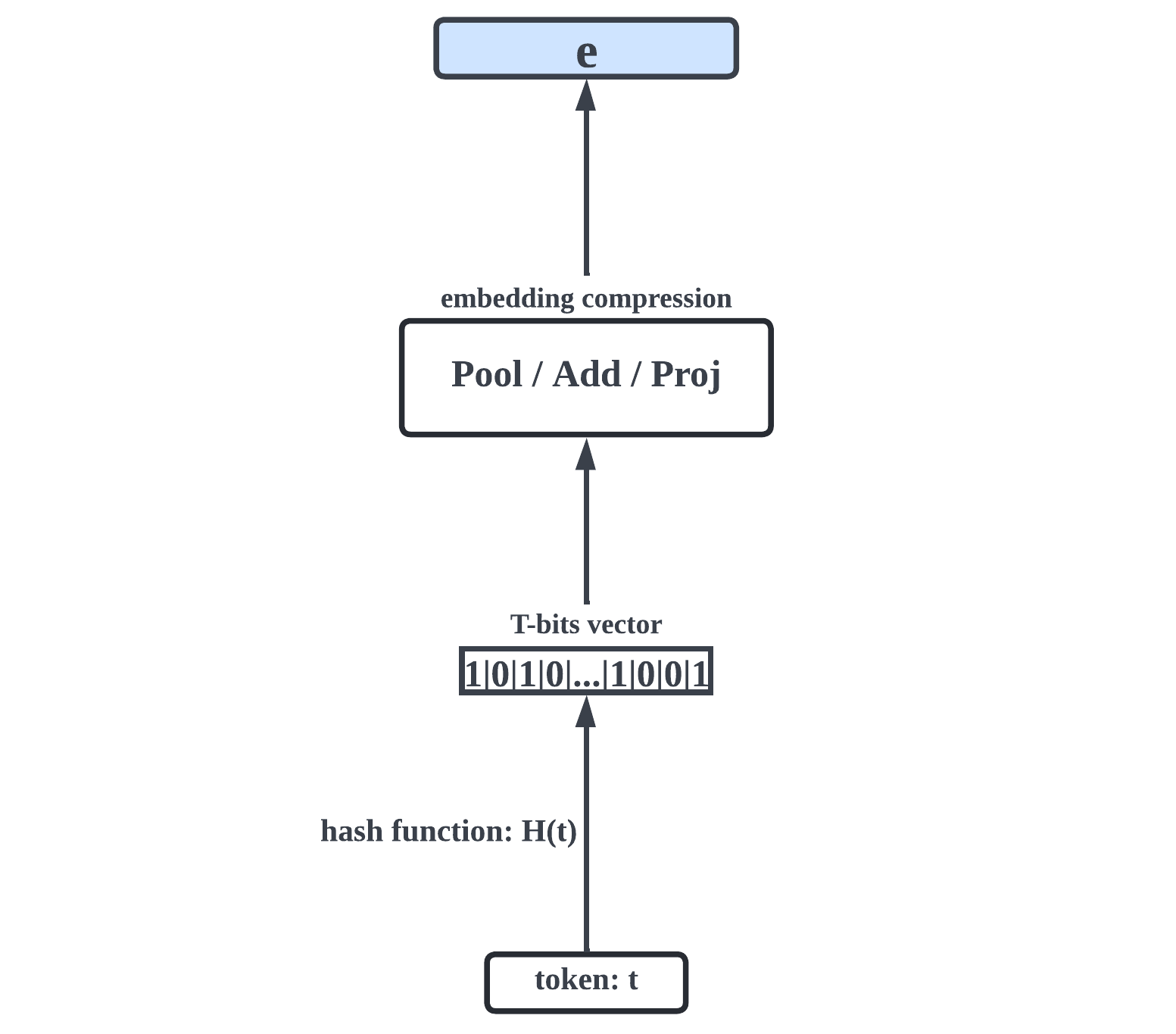}
    \caption{Compressing the embedding space.}
    \label{fig:fig-hformer-comp}
\end{figure}

We also propose three embedding compression approaches that allow an even smaller number of parameters to represent token embeddings and support unlimited tokens (i.e. very large $|V|$) without forcing a large number of tokens to share the same embedding. For this purpose, we first use a hash function $\mathcal{H}$ to map each token $t$ into a $T$-bit value $b$, $b \in [0,2^T)$. Then, we pass $b$ through a transformation procedure to generate the corresponding embedding (to facilitate computation, we cast $b$ into a $T$-bit vector $\mathbf{\tau}$). This way tokens with different values $b$ will be assigned to a different embedding by keeping the number of parameters relatively small. \autoref{fig:fig-hformer-comp} shows an overview of this method.

\begin{figure}[!h]
    \centering
    \includegraphics[width=0.48\textwidth]{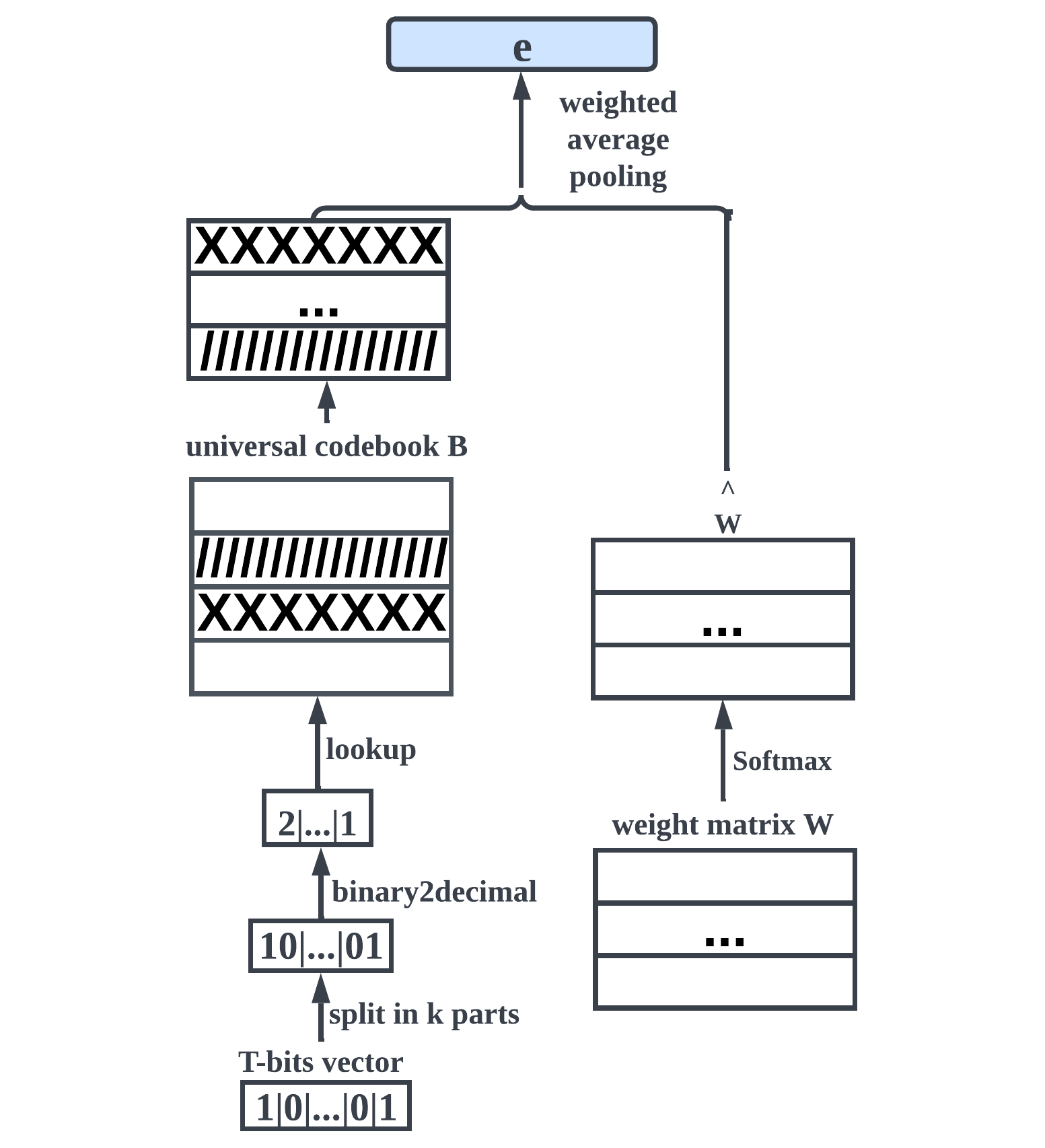}
    \caption{\textsc{HashFormer}-Pool.}
    \label{fig:fig-hformer-pool}
\end{figure}

\paragraph{Pooling Approach (Pool)} 
Inspired by \citet{svenstrup2017hash} and \citet{prakash-etal-2020-compressing}, we first create a universal learnable codebook, which is a matrix denoted as $\mathbf{B}\in\mathbb{R}^{2^k\times d}$. Then, we split the hash bit vector $\tau$ in $k$ successive bits without overlap to obtain $\lceil\frac{T}{k}\rceil$ binary values. We then cast these binary values into an integer value representing a codeword. Hence, each token is represented by a vector $\mathbf{c}\in\mathbb{R}^d$ with elements $c_j\in[0,2^k)$. For example, given $k=4$ and a 12-bits vector [1,0,1,0,0,1,0,0,0,0,0,1], 4-bit parts are treated as separate binary codewords [1010,0100,0001] then transformed into their decimal format codebook [10,4,1]. We construct the embedding $\mathbf{e}\in\mathbb{R}^d$ for each token by looking up the decimal codebook and extracting $\lceil\frac{T}{k}\rceil$ vectors corresponding to its $\lceil\frac{T}{k}\rceil$ codewords. We then apply a weighted average pooling on them using a softmax function:
\begin{subequations}
    \begin{align}
      \hat{\textbf{W}_{j}} &= \frac{\exp \textbf{W}_{j}}{\sum_{l=1}^{\lceil\frac{T}{k}\rceil} \exp \textbf{W}_{l}}, j=1,..,\lceil\tfrac{T}{k}\rceil \label{eq:1}\\
      \mathbf{e} &= \sum_{j=1}^{\lceil\frac{T}{k}\rceil}[\mathbf{B}_{\mathbf{c}}\odot \hat{\textbf{W}}]_{j} \label{eq:2}
    \end{align}
\end{subequations}
where $\mathbf{W}\in\mathbb{R}^{\lceil\frac{T}{k}\rceil\times d}$ is a learnable weight matrix as well as the codebook $\mathbf{B}$. The total number of parameters required for this pooling transformation is $(\lceil\frac{T}{k}\rceil+2^k)\times d$. This can be much smaller than the $N \times d$ parameters required for standard PLMs that use a one-to-one mapping between tokens and embeddings, where $N=|V|\gg(\lceil\frac{T}{k}\rceil+2^k)$. \autoref{fig:fig-hformer-pool} shows the overview of the Pool process.

\paragraph{Additive Approach (Add)} 
Different to the Pool method that uses a universal codebook, we create $T$ different codebooks $\{\textbf{B}^1,\textbf{B}^2,...,\textbf{B}^T\}$, each containing two learnable embedding vectors corresponding to codewords $0$ and $1$ respectively. We get a $T$-bits vector $\mathbf{\tau} \in \{0,1\}^{T}$ for each token, where each element in the vector $\mathbf{\tau}$ is treated as a codeword. We look up each codeword in their corresponding codebook to obtain $T$ vectors and add up them to compute the token embedding $\mathbf{e}$:
\begin{equation}
        \mathbf{e} = \sum^{T}_{j=1}\textbf{B}^j_{\mathbf{\tau}_j}\bigg/\gamma\label{eq:3}
\end{equation}
where $\mathbf{B}^j \in \mathbb{R}^{2\times d}, j=1,..,T$, $\gamma$ is the scaling factor.\footnote{Instead of averaging ($\gamma=T$), we set $\gamma=\sqrt{T}$ which we found to perform better in early experimentation.} Hence, the total number of parameters the additive transformation approach requires is $2\times T\times d$. Similar to the Pool approach, the number of parameters required is smaller than the vocabulary size: $2\times T\times d \ll N=|V|$.

\begin{figure}[!h]
    \centering
    \includegraphics[width=0.48\textwidth]{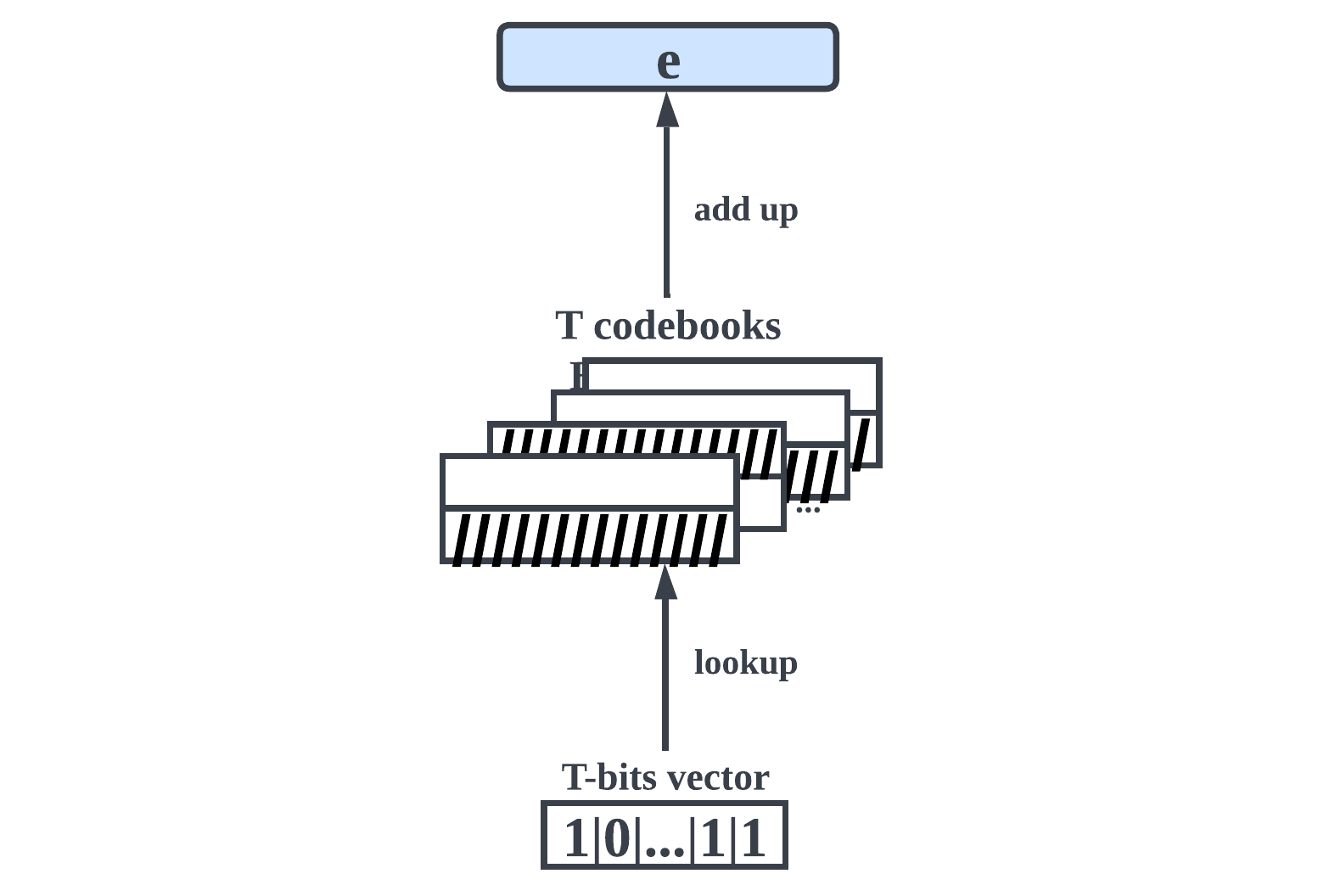}
    \caption{\textsc{HashFormer}-Add.}
    \label{fig:fig-hformer-add}
\end{figure}

\begin{figure}[!htp]
    \centering
    \includegraphics[width=0.48\textwidth]{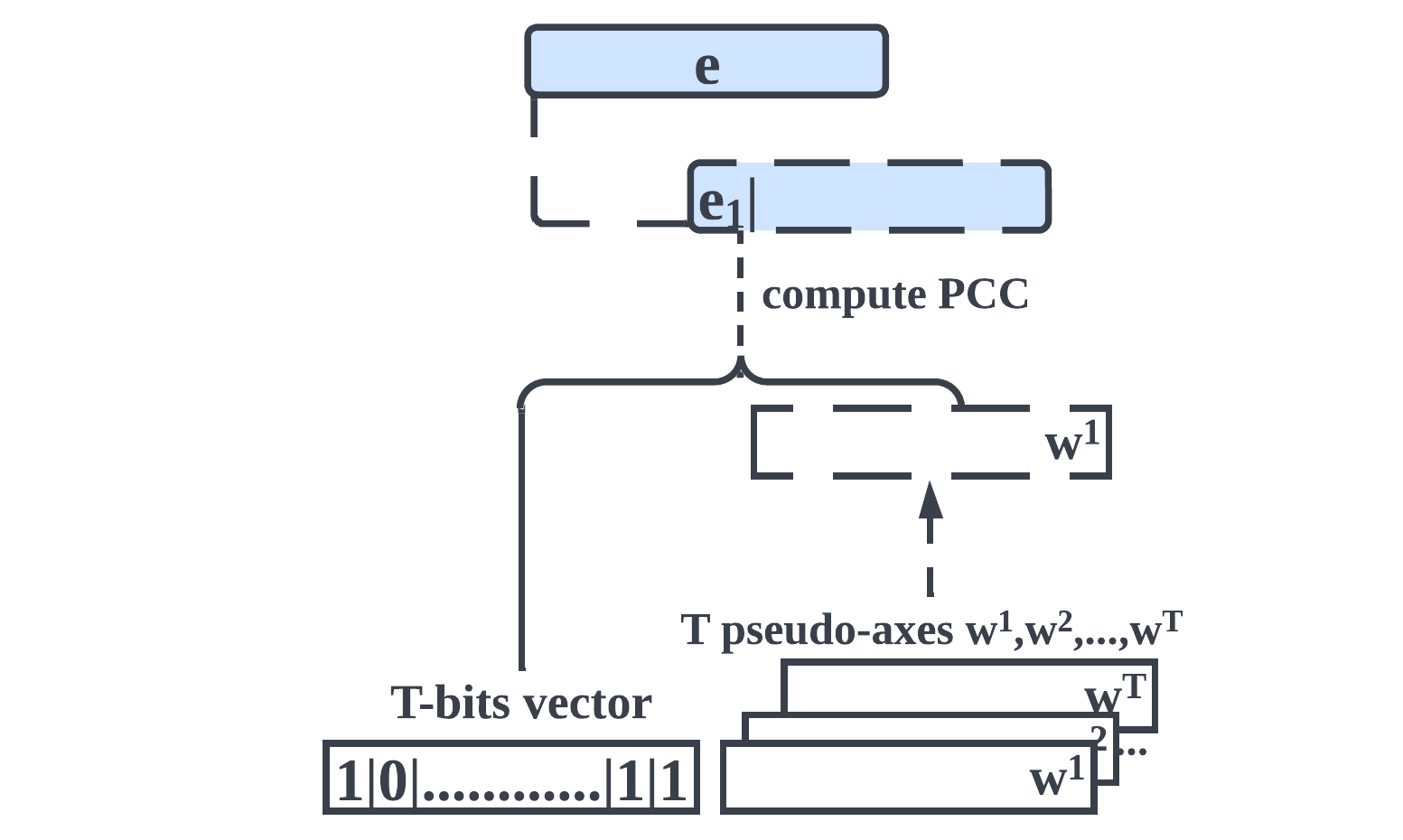}
    \caption{\textsc{HashFormer}-Proj.}
    \label{fig:fig-hformer-proj}
\end{figure}

\paragraph{Projection Approach (Proj)} 
Finally, we propose a new simpler approach compared to Pool and Add. We create $T$ learnable random initialized vectors as $T$ pseudo-axes to trace the orientation of each $T$-bits vector $\mathbf{\tau}$ corresponding to the token $t$. Given a token bit vector $\mathbf{\tau}$, the $j$\ts{th} element in the embedding $\mathbf{e}$  is computed as the Pearson's correlation coefficient (PCC) between $\mathbf{\tau}$ and the learnable vector $\mathbf{w}^{j}$ corresponding to $j$.
\begin{equation}
  \begin{aligned}
    \mathbf{e}_{j} &= \frac{\langle \mathbf\tau-\bar{\tau}, \mathbf{w^{j}}-\bar{\mathbf{w^{j}}} \rangle}{\Vert \tau-\bar{\tau} \Vert \centerdot \Vert \mathbf{w^{j}}-\bar{\mathbf{w^{j}}} \Vert}, \quad j=1,...,d \label{eq:4} \\
    \mathbf{e} &= (e_1, ..., e_{d})
  \end{aligned}
\end{equation}
$\mathbf{w}^{j}\in\mathbb{R}^{d}, j=1,..,T$, hence, the total number of parameters the projection transformation approach requires is only $T\times d \ll N=|V|$. \autoref{fig:fig-hformer-proj} depicts an overview of our \textsc{HashFormer}-Proj model.

\subsection{Hashing for Compressed Embeddings}
Similar to the embedding based \textsc{HashFormers}-Emb, our embedding compression-based models also consider the same two hash approaches (MD and LSH) for generating the $T$-bit vector of each token.

\paragraph{MD5:} We directly map the tokens to its 128-bits output $b$ with a universal secret key.

\paragraph{LSH:} We repeat the same morphological feature extraction step to obtain a feature vector $\mathbf{x}$ corresponding to each token $t$. However, rather than using $2^T$ random hyperplanes that require storing vectors of size $\mathbb{R}^{2^T}$, we simply use $T$ random hyperplanes similar to ~\citet{ravi2019efficient,sankar-etal-2021-proformer}. Each bit in $b$ represents which side of the corresponding hyperplane $\mathbf{r}\in\mathbb{R}^d$ the feature vector $\mathbf{x}$ is located: $b_j=\sgn(\sgn(\mathbf{x}\cdot\mathbf{r^i})+1), j=1,...,T$. This allows an on-the-fly computation without storing any vector~\citep{ravi2019efficient}.

\subsection{Pre-training Objective}
Since our models support an arbitrary number of unique tokens, it is intractable to use a standard Masked Language Modeling~\citep{devlin-etal-2019-bert} pre-training objective. We opted using \textsc{Shuffle + Random} (S+R), a computationally efficient three-way classification objective introduced by \citet{yamaguchi-etal-2021-frustratingly} for predicting whether tokens in the input have been shuffled, replaced with random tokens or remain intact.

\section{Experimental Setup}

\subsection{Baseline Models}
We compare \textsc{HashFormers} against the following baselines: (i) a \textsc{BERT}-base model~\citep{devlin-etal-2019-bert} with BPE tokenization and an MLM objective (\textsc{BERT}-MLM); (ii) another \textsc{BERT}-base model with BPE tokenization and a Shuffle+Random objective (\textsc{BERT}-S+R); (iii) \textsc{Canine}-C\footnote{We use the off-the-shelf \textsc{Canine}-C from \url{https://huggingface.co/google/canine-c}.}~\citep{clark-et-al2022-canine}
 a vocabulary-free pre-trained PLM on Unicode character sequences; (iv) \textsc{ProFormer}\footnote{ProFormer is not open-source, hence we have re-implemented it following the description of the model in the paper.}~\citep{sankar-etal-2021-proformer} a vocabulary-free LSH projection based transformer model with two encoder layers that is not pre-trained but only trained from scratch on the task at hand.

\subsection{Implementation Details}

\paragraph{Model Architecture}
Following the architecture of \textsc{BERT}-base, we use 12 transformer layers, an embedding size of 768 and a maximum sequence length of 512.\footnote{We note that the transformer encoder could easily be replaced with any other encoder.} For \textsc{HashFormers-LSH}, we set $T=128$ to make it comparable to \textsc{HashFormers-MD}, as MD5 produces a 128-bit hash value. For \textsc{HashFormer-MD}-Pool and \textsc{HashFormer-LSH}-Pool, we choose $k=10$ to keep the number of total parameters for the embeddings relatively small. We also experiment with two sizes of the embedding matrix of \textsc{HashFormers}-Emb for MD and LSH hashing. The first uses an embedding matrix of 50K, the same number of embedding parameters as \textsc{BERT}-base, while the second uses 1K which is closer to the size of the smaller Pool, Add and Proj models.

\paragraph{Hyperparameters}
Hyperparameter selection details are in \autoref{sec:hyperparameters}.

\paragraph{Pre-training}
We pre-train all \textsc{HashFormers}, \textsc{BERT}-MLM and \textsc{BERT}-S+R on the English Wikipedia and BookCorpus ~\citep{zhu2015aligning} from HuggingFace ~\citep{lhoest-etal-2021-datasets} for up to 500k steps with a batch size of 128. For our \textsc{HashFormer} models, we use white space tokenization resulting into a vocabulary of 11,890,081 unique tokens. For \textsc{BERT}-MLM and \textsc{BERT}-S+R, we use a 50,000 BPE vocabulary~\cite{liu2019roberta}.

\paragraph{Hardware}
For pre-training, we use eight NVIDIA Tesla V100 GPUs. For fine-tuning on downstream tasks, we use one NVIDIA Tesla V100 GPU.

\subsection{Predictive Performance Evaluation} \label{sec: Evaluation}
We evaluate all models on \textsc{Glue}~\citep{wang-etal-2018-glue} benchmark. We report matched accuracy for MNLI, Matthews correlation for CoLA, Spearman correlation for STS, F1 score for QQP and accuracy for all other tasks.

\renewcommand*{\arraystretch}{1.1}
\begin{table*}[!htp]
\begin{center}
\small
\resizebox{\linewidth}{!}{%
\begin{tabular}{lccccccccc|c}
\toprule

Model & Token & MNLI & QNLI & QQP & RTE & SST & MRPC & CoLA & STS & Avg.\\ \midrule
BERT-MLM & subword &  \bf{81.9} & \bf{88.9} & 86.7 & 60.4 & \bf{92.0} & 85.7 & 54.5 & 86.0 & 79.5(0.4)\\
BERT-S+R & subword &  79.9 & 88.7 & 86.7 & 64.6 & 88.6 & 85.6 & 55.6 & \bf{86.8} & 79.6(0.3)\\ 
CANINE-C & unicode & 77.7 & 87.6 & 82.8 & 62.0 & 85.7 & 81.4 & 2.3 & 83.9 & 70.4(1.3)\\
ProFormer & word & 45.2 & 59.1 & 71.4 & 53.9 & 82.1 & 71.2 & 9.7 & 22.1 & 51.8(0.5)\\
\midrule
\multicolumn{3}{l}{HashFormers-MD (Ours)} & \multicolumn{8}{c}{} \\ 
\enskip\enskip Emb (50K) & word & 79.6 & 88.4 & \bf{86.9} & \bf{66.4} &	88.0 & \bf{86.8} & 57.3 & 86.1 & \bf{79.9}(0.3)\\
\enskip\enskip Emb (1K) & word & 67.9 &	80.5 & 81.0 & 55.8 & 72.9 & 78.4 & 19.0 & 79.0 & 66.8(0.9)\\
\enskip\enskip Pool & word & 75.6 & 84.9 & 84.9 & 59.7 & 86.7 & 82.7 & 45.7 & 82.0 & 75.3(0.2)\\
\enskip\enskip Add & word  & 76.2 & 86.3 & 85.2 & 60.2 & 86.6 & 81.9 & 47.4 & 82.2 & 75.7(0.5)\\
\enskip\enskip Proj & word  & 76.0 & 85.8 & 84.8 & 60.9 & 87.3 & 83.0 & 45.9 & 82.1 & 75.7(0.3)\\
\midrule

\multicolumn{3}{l}{HashFormers-LSH (Ours)} & \multicolumn{8}{c}{} \\ 
\enskip\enskip Emb (50K) & word  & 76.1 & 86.5 & 85.5 &	65.5 & 83.6 & 84.2 & 42.7 & 83.7 & 76.0(0.3)\\
\enskip\enskip Emb (1K) & word & 65.6 & 80.1 & 80.0 & 56.4 & 71.3 & 78.1 & 5.2 & 76.9 & 64.2(0.8)\\
\enskip\enskip Pool & word  & 78.0 & 87.7 & 86.4 & 65.6 & 88.1 & 84.2 & 55.3 & 85.6 & 78.9(0.3)\\
\enskip\enskip Add & word  & 78.6 & 88.2 & 86.0 & 63.1 & 88.0 & 84.0 & \bf{57.7} & 85.9 & 78.9(0.2)\\
\enskip\enskip Proj & word  & 79.2 & 88.7 & 86.5 & 63.4 & 88.9 & 84.6 & 56.2 & 85.5 & 79.1(0.3)\\

\bottomrule
\end{tabular}%
}
\caption{Results on \textsc{Glue} dev sets with standard deviations over three runs in parentheses. \textbf{Bold} values denote best performing method in each task. } 

\label{table:base_result}
\end{center}
\end{table*}

\subsection{Efficiency Evaluation}\label{sec:Efficiency}
Furthermore, we use the following metrics to measure and compare the memory and computational efficiency of \textsc{HashFormers} and the baselines.

\paragraph{Memory Efficiency Metrics}

We define the three memory efficiency metrics together with a performance retention metric to use it as a point of reference:

\begin{itemize}
\item \textbf{Performance Retention Ratio:} We compute the ratio between the predictive performance of our target model compared to a baseline model performance. A higher PRR indicates better performance.
\begin{equation}
\mathbf{PRR} = \frac{\mathbf{score}_{model}}{\mathbf{score}_{baseline}}
\end{equation}

\item\textbf{Parameters Compression Ratio (All):}
We compute use the ratio between the total number of parameters of our target model and that of the baseline to measure the memory efficiency of the target model compared to the baseline. A higher $\text{PCR}_\text{All}$ score indicates better memory efficiency for the entire model.
\begin{equation}
\mathbf{PCR_{All}} = 1-\frac{\mathbf{\#}_{Model Params}}{\mathbf{\#}_{baseline Params}}
\end{equation}

\item\textbf{Parameters Compression Ratio (Emb):}
We also use the ratio between the number of parameters required by a target model for representing embeddings and that of the baseline. A higher $\text{PCR}_\text{Emb}$ score indicates better memory efficiency for the embedding representation.
\begin{equation}
\mathbf{PCR_{Emb}} = 1-\frac{\mathbf{\#}_{Model Emb Params}}{\mathbf{\#}_{baseline Emb Params}}
\end{equation}

\item\textbf{Proportion of Embedding Parameters:}
We also use the proportion of parameters of embeddings out of the total parameters of each model to show the memory footprint of the embedding space on each model.
\begin{equation}
\mathbf{PoEP} = \frac{\mathbf{\#}_{Emb Params}}{\mathbf{\#}_{Total Params}}
\end{equation}
Ideally, we expect a smaller PoEP, indicating that the embedding parameters occupy as little memory as possible out of the total number of parameters of a model.

\end{itemize}

For number of parameters calculations, please see \autoref{sec:Parameters Counts}.

\paragraph{Computational Efficiency Metrics}
We also measure the computational efficiency for pre-training (PT) and inference (Infer). Each pre-training step is defined as a forward pass and a backward pass. The inference is defined by a single forward pass.
\begin{itemize}
\item\textbf{Time per Sample (Time)}
This measures the average time of a sample completing a PT or Infer step. It is measured in milliseconds (ms)/sample. Lower PT and Infer time indicates a more computational efficient model.

\item\textbf{Speed-up Rate}
We finally measure the model's computation speed-up rate against a baseline. It is defined as:
\begin{equation}
\mathbf{Speed\text{-}upRate} = 1\bigg/\frac{\mathbf{Time}_{model}}{\mathbf{Time}_{baseline}}
\end{equation}

\end{itemize}

\section{Results} \label{sec:Results}
\subsection{Predictive Performance Comparison}
\autoref{table:base_result} presents results on \textsc{Glue} for our \textsc{HashFormers} models and all baselines.
We first observe that both the performance of our \textsc{HashFormers}-Emb models (MD and LSH) are comparable to the two \textsc{BERT} variants (MLM and S+R) and CANINE-C on average \textsc{Glue} score (79.9 and 76.0 vs. 79.5, 79.6 and 70.4 respectively). Surprisingly, the more sophisticated \textsc{HashFormer-LSH}-Emb that takes morphological similarity of tokens into account does not outperform \textsc{HashFormer-MD}-Emb that uses a random hashing. We believe that \textsc{HashFormer-MD} generally outperforms \textsc{HashFormer-LSH} mainly due to its ability to map morphologically similar tokens to different vectors. This way it can distinguish tenses etc.. On the other hand, \textsc{HashFormer-LSH} confuses words with high morphological similarity (e.g. play, played) because it assigns them to the same embedding.

However, LSH contributes to the performance improvement of smaller \textsc{HashFormers} with compressed embedding spaces compared to their MD variants, i.e. Add (78.9 vs. 75.3), Add (78.9 vs. 75.7) and Proj (79.1 vs. 75.7). The best performing compressed \textsc{HashFormer-LSH}-Proj model obtains 79.1 average \textsc{Glue} score, which is only 0.4 lower than the BERT baselines. Reducing the number of embedding vectors in Emb (1K) models is detrimental to performance and leads to drastic drops between 11.8\% and 13.1\%. This indicates that the model size plays a more important role than the choice of tokenization approach (i.e. white space or BPE) or the vocabulary size (i.e. 12M vs. 50K). At the same time, comparing to Emb, the Pool, Add and Proj approaches do not suffer from predictive accuracy degradation, i.e. 0.4-4.2\%.

All our \textsc{HashFormers} show clear advantages comparing to the LSH based \textsc{ProFormer} which is not pre-trained across the majority of tasks (i.e. MNLI, QNLI, QQP, MRPC, CoLA and STS). Although \textsc{ProFormer} shows that for a relatively simpler sentiment analysis task (SST), pre-training might not be necessary.

\subsection{Memory Efficiency Comparison}

\renewcommand*{\arraystretch}{1.1}
\begin{table*}[!t]
\begin{center}
\resizebox{\linewidth}{!}{%
\strutlongstacks{T}
\begin{tabular}{lcccccccc|c|rrrrr}
\toprule

Emb. & MNLI & QNLI & QQP & RTE & SST & MRPC & CoLA & STS & \textsc{Glue} Avg. & \makecell{ \#Total \\Params} & \makecell{ \#Emb \\Params} & \makecell{PCR \\(All)} & \makecell{PCR \\(Emb)} & \makecell{PoEP}\\ \midrule
CANINE-C & 94.9 & 98.5 & 95.5 & 102.6 & 93.2 & 95.0	& 4.2 & 97.6 & 88.6 & 121.0M & 12.3M & 2.9 & 68.1 & 10.2 \\ 
ProFormer & 55.2 & 66.5 & 82.4 & 89.2 & 89.2 & 83.1 & 17.8 & 25.7 & 65.2 & \bf{15.1M} & 322.6K & \bf{87.9} & 99.2 & 2.1 \\\midrule
\multicolumn{15}{l}{HashFormers-MD (Ours)}\\
\enskip Emb (50K) & \bf{97.2} & 99.4 & \bf{100.2} & \bf{109.9} & 95.7 & \bf{101.3} & 105.1 & \bf{100.1} & \bf{100.5} & 124.6M & 38.6M & 0.0 & 0.0 & 31.0 \\
\enskip Emb (1K) & 82.9 & 90.6 & 93.4 & 92.4 & 79.2	& 91.5 & 34.9 & 91.9 & 84.0 & 86.8M & 797.2K & 30.3 & 97.9 & 1.0 \\
\enskip Pool & 92.3	& 95.5 & 97.9 & 98.8 & 94.2	& 96.5 & 83.9 & 95.3 & 94.7 & 86.8M & 797.2K & 30.3 & 97.9 & 1.0 \\
\enskip Add & 93.0 & 97.1 & 98.3 & 99.7	& 94.1 & 95.6 & 87.0 & 95.6	& 95.2 & 86.2M & 197.4K & 30.8 & 99.5 & 0.2 \\
\enskip Proj & 92.8	& 96.5 & 97.8 & 100.8 & 94.9 & 96.8	& 84.2 & 95.5 & 95.2 & 86.1M & \textbf{99.1K} & 30.9 & \textbf{99.7} & \textbf{0.1} \\
\midrule

\multicolumn{15}{l}{HashFormers-LSH (Ours)}\\
\enskip Emb (50K) & 92.9 & 97.3	& 98.6 & 108.4 & 90.9 & 98.2 & 78.3	& 97.3 & 95.6 & 124.6M & 38.6M & 0.0 & 0.0 & 31.0 \\
\enskip Emb (1K) & 80.1	& 90.1 & 92.3 & 93.4 & 77.5	& 91.1 & 9.5 & 89.4	& 80.8 & 86.8M & 797.2K & 30.3 & 97.9 & 1.0 \\
\enskip Pool & 95.2	& 98.7 & 99.7 & 108.6 & 95.8 & 98.2	& 101.5	& 99.5 & 99.2 & 86.8M & 797.2K & 30.3 & 97.9 & 1.0 \\
\enskip Add & 96.0 & 99.2 & 99.2 & 104.5 & 95.7 & 98.0 & \bf{105.9} & 99.9 & 99.2 & 86.2M & 197.4K & 30.8 & 99.5 & 0.2 \\
\enskip Proj & 96.7	& \bf{99.8}	& 99.8 & 105.0 & \bf{96.6} & 98.7 & 103.1 & 99.4	& 99.5 & 86.1M & \textbf{99.1K} & 30.9 & \textbf{99.7} & \textbf{0.1} \\

\bottomrule
\end{tabular}%
}
\caption{Memory efficiency metrics and performance retention (\%) on \textsc{Glue} for \textsc{HashFormer} models, \textsc{CANINE}-C and ProFormer using \textsc{BERT}-MLM as a baseline.} 

\label{table:save_result}
\end{center}
\end{table*}

\autoref{table:save_result} shows the results on memory efficiency and performance retention (\%) on GLUE using BERT-MLM as a baseline.
Notably, Pool, Add and Proj models provide large compression to the total number of embeddings parameters compared to Emb as well as \textsc{Canine}-C and \textsc{BERT} variants. This is approximately a 30\% $\text{PCR}_\text{All}$ and 97-99\% $\text{PCR}_\text{Emb}$ compared to \textsc{BERT}. These models also achieve very high performance retention (from 94.7\% to 99.5\%) which highlights their efficiency. In one case, \textsc{HashFormer-LSH}-Add outperforms the BERT-MLM baseline on CoLA with a retention ratio of 105.9\% using only 197.4K parameters for token embeddings.
 
Proj variants, the smallest of \textsc{HashFormers} achieve the highest performance retention (95.2\% with MD, 99.5\% with LSH) compared to Pool (94.7\% with MD, 99.2\% with LSH) and Add (95.2\% with MD, 99.2\% with LSH). 
Overall, they only have a negligible drop in performance retention (0.5\%) while they are extremely more memory efficient. Proj uses a substantially smaller number of embedding parameters (99.1K) compared to \textsc{Canine}-C and \textsc{BERT} variants (i.e., 12.3M and 38.6M respectively). In general, Pool, Add and Proj models lead to a 30\% reduction in the total number of parameters (around 30.0M) compared to the baseline model and make their embedding footprint minimal, i.e. 0.1-1\% PoEP. On the other hand, \textsc{Canine}-C has a larger embedding footprint (10.2\% PoEP) but with similar or smaller performance retention compared to \textsc{HashFormers}.

\renewcommand*{\arraystretch}{1.1}
\begin{table}[!t]
\begin{center}
\small
\resizebox{\linewidth}{!}{%
\begin{tabular}{lcccc}
\toprule

\makecell{Model}  & \makecell{PT \\ Time \\ (ms/samp)} & \makecell{PT \\ Speed-up \\ Rate} & \makecell{Infer \\ Time \\ (ms/samp)} & \makecell{Infer \\ Speed-up \\ Rate} \\\midrule
\multicolumn{5}{l}{\enskip \textsc{BERT}} \\
\enskip\enskip -MLM & 24.9 & 1.0 & 4.6 & 1.0 \\
\enskip\enskip -S+R & 11.6 & 2.1x & 4.6 & 1.0x \\ \midrule
\enskip \textsc{Canine}-C & - & - & 6.9 & 0.6x \\\midrule
\multicolumn{5}{l}{\enskip \textsc{HashFormers} (Ours)} \\
\enskip\enskip -Emb & 11.6 & 2.1x & 2.0\textasciitilde4.6 & 1.0x\textasciitilde2.4x \\
\enskip\enskip -Pool & 12.0 & 2.1x & 2.0\textasciitilde4.6 & 1.0x\textasciitilde2.3x \\
\enskip\enskip -Add & 11.7 & 2.1x & 2.0\textasciitilde4.6 & 1.0x\textasciitilde2.4x \\
\enskip\enskip -Proj & 10.6 & 2.4x & 1.8\textasciitilde4.6 & 1.0x\textasciitilde2.6x \\

\bottomrule
\end{tabular}%
}
\caption{Results on pre-training speed and inference speed under different embedding compression strategies. We use \textsc{BERT}-MLM as the baseline model. The sequence length is fixed to 512 for pre-training. For inference, sequence length is equal to the length of the longest sequence in the batch.} 

\label{table:speed_result}
\end{center}
\end{table}

\textsc{HashFormers}-Emb have an embedding matrix of equal size (i.e. 50K embeddings) as \textsc{BERT}. However, \textsc{BERT} only supports a vocabulary of 50K tokens, while \textsc{HashFormers}-Emb supports an unlimited vocabulary, e.g. 12M unique tokens in our pre-training corpora. Using a smaller embedding matrix (i.e. 1K), the performance retention drops 20\%\textasciitilde26\%. Despite the fact that \textsc{HashFormers}-Emb (1K) has a similar number of embedding parameters as the embedding compression approaches (i.e. Pool, Add, Proj), it falls far behind those models, i.e. between 8.5\% and 14.3\% for both MD and LSH variants. This demonstrates the effectiveness of our proposed embedding compression approaches.

Although, the more lightweight ProFormer with only two transformer layers consists of 15.1M parameters in total (approximately a 87.9\% $\text{PCR}_\text{All}$), its performance\footnote{The predictive performance of ProFormer does not improve, even if we train it for four times more epochs (20 epochs). We report the results when trained for a maximum of five epochs.} fall far behind our worst \textsc{HashFormer-MD}-Pool with a difference of 29.5\% PRR on \textsc{Glue} Avg. score. Nevertheless, ProFormer requires more bits for hashing the tokens, resulting in more parameters for representing token embeddings (322.6K) comparing to \textsc{HashFormers}-Add and \textsc{HashFormers}-Proj (197.4K and 99.1K). Such memory efficiency gains substantially sacrifice model's predictive performance.


\subsection{Computational Efficiency Comparison}
\autoref{table:speed_result} shows the pre-training (PT) and inference (Infer) time per sample for \textsc{HashFormers}, \textsc{Canine}-C, \textsc{BERT}-S+R using \textsc{BERT}-MLM as a baseline for reference. 
We note that \textsc{HashFormers} have comparable pre-training training time (PT) to the fastest BERT model (BERT-S+R). This highlights that the complexity of the pre-training objective is more important than the size of the embedding matrix for improving computational efficiency for pre-training. 

During inference, we observe that the speed-up obtained by \textsc{HashFormers} is up to 2.6x compared to both BERT models. However, this is due to the tokenization approach. \textsc{HashFormers} operate on the word level, so the sequence length of the input data is smaller, leading to inference speed-ups.  
Finally, we observe that \textsc{Canine}-C has a slower inference time compared to both \textsc{BERT} models and \textsc{HashFormers}. This might be due to its relatively more complex approach for processing the long Unicode character input sequence. 

\section{Conclusions}
We have proposed \textsc{HashFormers}, a family of vocabulary-independent hashing-based pre-trained transformers. 
We have empirically demonstrated that our models are computationally cheaper and more memory efficient compared to standard pre-trained transformers, requiring only a fraction of their parameters to represent token embeddings. \textsc{HashFormer-LSH}-Proj variant needs 99.1K parameters for representing the embeddings compared to millions of parameters required by state-of-the-art models with only a negligible performance degradation.
For future work, we plan to explore multilingual pre-training with \textsc{HashFormers} and explore their ability in encoding linguistic properties~\cite{alajrami-aletras-2022-pre}.

\section*{Limitations}
We experiment only using English data to make comparisons with previous work easier. For languages without explicit white spaces (e.g. Chinese and Japanese), our methods can be applied with different tokenization techniques, e.g. using a fixed length window of characters.

\section*{Acknowledgments}
This project made use of time on Tier 2 HPC facility JADE2, funded by EPSRC (EP/T022205/1). We would like to thank Miles Williams and the anonymous reviewers for their invaluable feedback.



\bibliography{anthology,custom}
\bibliographystyle{acl_natbib}

\clearpage

\appendix

\section{Hyperparameters}\label{sec:hyperparameters}
The hyperparameters used in pre-training are listed in \autoref{table:hyperparams_pretraining}.

\renewcommand*{\arraystretch}{1.0}
\begin{table}[!h]
\begin{center}
\small
\resizebox{\linewidth}{!}{%
\begin{tabular}{ll}
\toprule
Hyperparameter & Pretraining
\\ \midrule
Maximum train epochs & 10 epochs \\
Batch size (per GPU) & 16 instances \\
Adam $\epsilon$ & 1e-8 \\
Adam $\beta_1$ & 0.9 \\
Adam $\beta_2$ & 0.9999 \\
Sequence length & 512 \\
Peak learning rate & 1e-4 for MLM, 5e-5 for others \\
Learning rate schedule & linear \\
Warmup steps & 10000 \\
Weight decay & 0.01 \\
Attention Dropout & 0.1 \\
Dropout & 0.1 \\

\bottomrule
\end{tabular}%
}
\caption{Details of hyperparameters used in pre-training.} 

\label{table:hyperparams_pretraining}
\end{center}
\end{table}

The hyperparameters used in fine-tuning are listed in \autoref{table:hyperparams_finetuning}.

\renewcommand*{\arraystretch}{1.0}
\begin{table}[!h]
\begin{center}
\small
\resizebox{\linewidth}{!}{%
\begin{tabular}{ll}
\toprule
Hyperparameter & Pretraining
\\ \midrule
Maximum train epochs & 5 epochs \\
Batch size (per GPU) & 32 instances \\
Adam $\epsilon$ & 1e-6 \\
Adam $\beta_1$ & 0.9 \\
Adam $\beta_2$ & 0.999 \\
Peak learning rate & 3e-5 \\
Learning rate schedule & cosine with hard restarts \\
Warmup steps & first 6\% steps\\
Weight decay & 0 \\
Attention Dropout & 0.1 \\
Dropout & 0.1 \\
Evaluation steps & 2455 for MNLI, 655 for QNLI, \\
& 2275 for QQP, 48 for RTE, \\
& 421 for SST, 69 for MRPC, \\
& 162 for CoLA and 108 for STS\\

\bottomrule
\end{tabular}%
}
\caption{Details of hyperparameters used in fine-tuning.} 

\label{table:hyperparams_finetuning}
\end{center}
\end{table}

\section{Model Parameter Counts}\label{sec:Parameters Counts}
We count the total number of parameters of each model on a binary classification task. This is computed by counting all learnable variables used for the task (including those in the classification head) without freezing any weights. For all \textsc{BERT} variants and our \textsc{HashFormers}, we adopt the same setting of \textsc{BERT}-Base by setting $Dim_{hidden} = 768$, $Dim_{intermediate}=3072$ with 12 hidden layers and 12 attention heads. For \textsc{CANINE}-C, we use the default base-sized model whose $Dim_{hidden} = 768$, $Dim_{intermediate}=3072$ and has 12 hidden layers and attention heads.

We only count the number of parameters which are used for retrieving or generating the embeddings of any tokens (excluding those special tokens e.g. <PAD>) and we also exclude those for position embeddings. Specifically, $\#_{[Model]EmbParams}$ are computed as the follow:

\begin{itemize}

    \item \textbf{\textsc{BERT} variants}:
    \begin{equation}
        \#_{BERT}=|V|\times d
    \end{equation}
    
    \item \textbf{\textsc{CANINE}-C}:
    \begin{equation}
        \#_{CANINE-C}=\#_{HashBuckets}\times d
    \end{equation}
    \textsc{CANINE-C} employs 16,000 hash buckets \citep{clark-et-al2022-canine}.
    
    \item \textbf{\textsc{ProFormer}}:
    \begin{equation}
        \#_{ProFormer}=\#_{LSHDigestSize}\times d
    \end{equation}
    \textsc{ProFormer} hashes each token into a 420-bit vector \citep{sankar-etal-2021-proformer}.
    
    \item \textbf{\textsc{HashFormers}-Emb}:
    \begin{equation}
        \#_{HashFormers}-Emb=N\times d
    \end{equation}
    
    \item \textbf{\textsc{HashFormers}-Pool}:
    \begin{equation}
        \#_{HashFormers}-Pool=(\lceil\frac{T}{k}+2^k\rceil)\times d
    \end{equation}
    
    \item \textbf{\textsc{HashFormers}-Add}:
    \begin{equation}
        \#_{HashFormers}-Add=2\times T\times d
    \end{equation}
    
    \item \textbf{\textsc{HashFormers}-Proj}:
    \begin{equation}
        \#_{HashFormers}-Proj=T\times d
    \end{equation}
    
\end{itemize}

\section{\textsc{HashFormers} with BPE Tokenization}\label{sec: hf_with_bpe}
\autoref{table:base_result_bpe} presents results on \textsc{Glue} for our \textsc{HashFomers} with BPE tokenization. In general, we observe that using BPE tokinization, the performance of \textsc{HashFomers} slightly drops.  

\renewcommand*{\arraystretch}{1.1}
\begin{table*}[!htp]
\begin{center}
\small
\resizebox{\linewidth}{!}{%
\begin{tabular}{lccccccccc|c}
\toprule

Model & Token & MNLI & QNLI & QQP & RTE & SST & MRPC & CoLA & STS & Avg.\\ \midrule
\multicolumn{5}{l}{HashFormers-MD} & \multicolumn{6}{c}{} \\ 
\enskip\enskip Emb (50K) & subword & 78.6 & 87.7 & 86.0 & 65.6 & 88.5 & 85.1 & 51.2 & 85.0 & 78.5(0.4)\\
\enskip\enskip Emb (50K) & word & 79.6 & 88.4 & 86.9 & 66.4 & 88 & 86.8 & 57.3 & 86.1 & 79.9(0.3)\\
\enskip\enskip Proj & subword & 74.6 & 84.8 & 83.7 & 58.7 & 85.5 & 80.7 & 44.6 & 80.1 & 74.1(0.5)\\
\enskip\enskip Proj & word & 76.0 & 85.8 & 84.8 & 60.9 & 87.3 & 83.0 & 45.9 & 82.1 & 75.7(0.3)\\
\midrule

\multicolumn{5}{l}{HashFormers-LSH} & \multicolumn{6}{c}{} \\ 
\enskip\enskip Emb (50K) & subword  & 62.6 & 80.2 & 80.8 & 59.3 & 71.3 & 80.2 & 18.3 & 75.5 & 66.0(0.2)\\
\enskip\enskip Emb (50K) & word  & 76.1 & 86.5 & 85.5 & 65.5 & 83.6	& 84.2 & 42.7 & 83.7 & 76.0(0.3)\\
\enskip\enskip Proj & subword  & 78.2 & 87.5 & 86.3 & 64.3 & 88.6 & 85.5 & 51.2 & 85.1 & 78.3(0.1)\\
\enskip\enskip Proj & word  & 79.2 & 88.7 & 86.5 & 63.4	& 88.9 & 84.6 & 56.2 & 85.5 & 79.1(0.3)\\

\bottomrule
\end{tabular}%
}
\caption{Results on \textsc{Glue} dev sets with standard deviations over three runs in parentheses using BPE tokenization.} 

\label{table:base_result_bpe}
\end{center}
\end{table*}

\end{document}